\documentclass{article}

\usepackage[nonatbib,final]{neurips_2024}
\usepackage[numbers]{natbib}

\usepackage[utf8]{inputenc} 
\usepackage[T1]{fontenc}    
\usepackage{hyperref}       
\usepackage{url}            
\usepackage{booktabs}       
\usepackage{amsfonts}       
\usepackage{nicefrac}       
\usepackage{microtype}      
\usepackage{xcolor}         
\usepackage{hyperref}
\usepackage{url}
\usepackage{graphicx}
\usepackage{subcaption}
\usepackage{soul}
\usepackage{algorithm}
\usepackage{algpseudocode}
\usepackage{tabularray}
\usepackage{wrapfig,lipsum,booktabs}
\usepackage{amsmath}

\def\sl#1{{\color{purple} SL: #1}}
\newcommand{\alg}{GAN-RL}

\title{Unsupervised Event Outlier Detection in Continuous Time}

\author{
Somjit Nath \\
McGill University, Mila \\
\texttt{\small{somjitnath@gmail.com}}
\And
Yik Chau Lui \\
Borealis AI \\
\texttt{\small{yikchau.y.lui@borealisai.com}}
\And
Siqi Liu \\ 
Borealis AI \\
\texttt{\small{siqi.x.liu@borealisai.com}}
}

\begin{document}

\maketitle

\begin{abstract}
   Event sequence data record the occurrences of events in continuous time. Event sequence forecasting based on temporal point processes (TPPs) has been extensively studied, but outlier or anomaly detection, especially without any supervision from humans, is still underexplored. In this work, we develop, to the best our knowledge, the first unsupervised outlier detection approach to detecting abnormal events. Our novel unsupervised outlier detection framework is based on ideas from generative adversarial networks (GANs) and reinforcement learning (RL). We train a ``generator'' that corrects outliers in the data with a ``discriminator'' that learns to discriminate the corrected data from the real data, which may contain outliers. A key insight is that if the generator made a mistake in the correction, it would generate anomalies that are different from the anomalies in the real data, so it serves as data augmentation for the discriminator learning. Different from typical GAN-based outlier detection approaches, our method employs the \emph{generator} to detect outliers in an \emph{online} manner. The experimental results show that our method can detect event outliers more accurately than the state-of-the-art approaches.
\end{abstract}

\section{Introduction}
Event sequence data are records of the occurrences of different events in continuous time, e.g., natural disasters in a country, or user actions when using an app. They can be represented as individual points on a timeline, with the location of the points indicating the time of the event occurrences. For event sequence data, forecasting and latent structure inference have been the focuses of most previous research. Methods based on Gaussian processes (e.g., \citep{rao2011gaussian,lloyd2015variational,lloyd2016latent,ding2018bayesian,liu2019nonparametric}), Hawkes processes (e.g., \citep{zhou2013learning,lee2016hawkes,xu2016learning,wang2016isotonic,kim2018markov}), and more recently deep neural networks (e.g., \citep{du2016recurrent,mei2017neural,xiao2017modeling,omi2019fully,zhang2020selfattentive,zuo2020transformer,xue2022hypro}), have been widely proposed and evaluated.

In contrast to forecasting, sometimes the event occurrences themselves can be unexpected, i.e. outliers. \citet{liu2021event} proposed a semi-supervised method, assuming access to clean data to train a model, to detect these outliers. Although their assumption is common in the literature, unsupervised methods without this assumption would have more practical value, since it is usually hard to get clean data without checking and preprocessing.

Inspired by Generative Adversarial Networks (GANs)~\citep{goodfellow2014Generative}, we propose to solve this problem by modelling a ``generator'' that tries to find and remove outliers and a ``discriminator'' that tries to distinguish the ``corrected'' data from the real data that can be either normal or abnormal. The key insight is that a generator can either correctly remove the outliers in the real data or incorrectly remove real points in the data. If the former is the case, it will be very difficult for the discriminator to separate the ``corrected'' data from the normal samples in the real data, which, by definition, constitute the majority of the data. Meanwhile, if the latter is the case, it will be relatively easy for the discriminator to separate. This intrinsic contrast between these two cases will be the source of feedback for both the generators and the discriminators to learn. Once learned, the ``generators'' can be used in an online manner to detect outliers in unseen event sequences. The discriminator can be trained by standard stochastic gradient descent algorithms. However, gradient descent-based optimization cannot be used for the generator because our case is non-differentiable. There are various ways for handling the non-differentiability, such as Gumbel-softmax \cite{jang2016categorical}, cooperative learning \cite{lu2019cot, lamprier2022generative} and policy gradient methods \citep{reinforce, a2c, ppo}. 
We chose the latter approach in this paper, due to its flexibility.

\section{Unsupervised Event Outlier Detection}
\begin{figure}[h]
    \centering
    \includegraphics[width=0.85\linewidth]{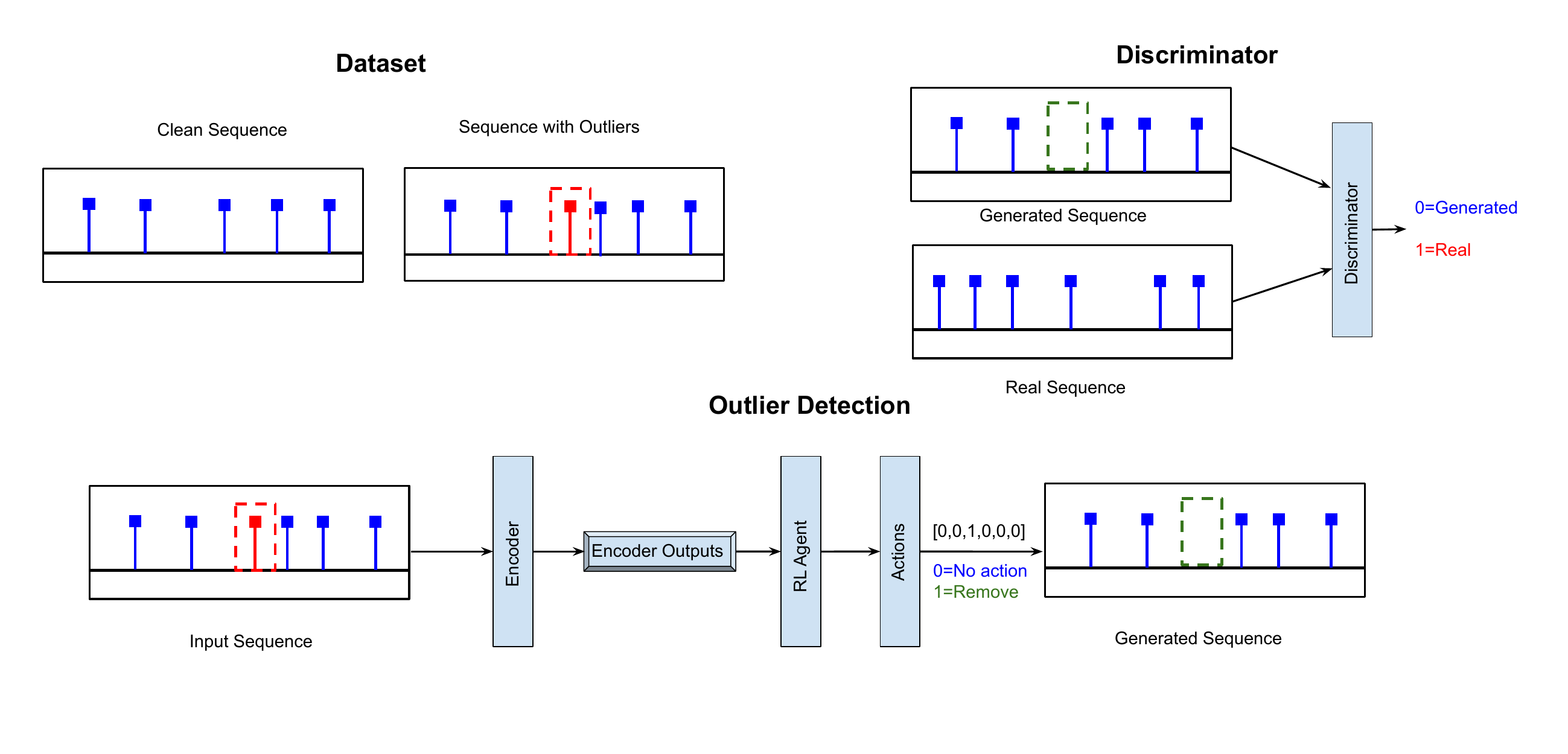}
    \caption{GAN + RL framework for unsupervised event outlier detection in continuous time.}
    \label{fig:method}
\end{figure}

\subsection{Problem Formulation}
An event sequence is defined as $S=\{t_n:t_n \in \mathcal{T}\}^{N}_{n=1}$ where $t_n$ is the time of the occurrence of event $n$, $N$ is the total number of events, and $\mathcal{T}$ denotes the entire time domain. We assume access to a dataset $\mathcal{D} = \{S_i\}_{i=1}^I$, consisting of $I$ event sequences, where some sequences might be corrupted in the form of the addition of abnormal points. Given an event point at $t_n$ in an event sequence $S$, the goal is to identify whether the event is an outlier or not, so the output would be a label, $y_n \in \{0, 1\}$, assigned to each event point.

In order to detect outliers in an unsupervised and online manner, we develop GAN-based approaches to train outlier detectors without any supervision. More specifically, we model a \textbf{generator that produces ``corrected" sequences} given input sequences from the potentially corrupted data and a \textbf{discriminator that distinguishes between the ``real'' and ``corrected'' sequences}.

We sample sequences from the potentially corrupted data as ``real'' sequences. Since by definition, outliers are supposed to be \emph{relatively rare} compared to normal data, we can reasonably assume that the majority of the sampled sequences are normal, and if the generator performed an incorrect ``correction'' on the data, e.g., removing a normal point instead of an outlier point from the sequence, then it would be possible for the discriminator to distinguish it from the sampled sequences. In the following sections, we describe our framework in detail. The final algorithm is in Appendix~\ref{sec:alg}.

\subsection{Encoder} \label{sec:encoder}
We use an encoder to summarize the information in the past for each sequence, as the occurrence of a new event can be influenced by the events that occurred before it. Since the event times are continuous and irregular, we use an architecture, the \textbf{continuous-time LSTMs (cLSTM)}~\citep{mei2017neural}, designed especially for continuous-time event sequences.
We can directly use the cLSTM outputs as inputs to the generator. Still, since we wish to detect and remove outliers, information about every point in the past can be crucial for determining the action, so we apply the \emph{attention} mechanism ~\citep{bahdanau2016neural} followed by \emph{layer normalization}~\citep{ba2016layer} to the latent outputs from the cLSTM with a causal mask to ensure that there is no information leak from the future. Therefore the learned outlier detector can be applied online. 
This entire architecture, consisting of the cLSTM and attention layer, forms the encoder as shown in Figure~\ref{fig:method}.

\subsection{Generator}\label{sec:gen}
For our problem, each generator is modelled as a Reinforcement Learning (RL) agent that tries to identify and remove outliers in continuous-time event sequences. Each sequence $S_i$, before being fed into the RL agent, is first passed through an encoder as we described in the previous section. The final encodings are treated by the RL agent as sequential states of the environment, whose goal is to identify and remove outlier points from the input sequence. Each sequence is treated as an episode, in which the agent makes decisions about making changes to the sequence. To achieve this, we use policy gradient methods, which are effective for parameterizing the optimal policy and exhibit good performance in these types of tasks~\citep{yoon2019data}. For our paper, we use the clip version of Proximal Policy Optimization (PPO) algorithm~\citep{ppo}.

The RL agent tries to identify and remove outlier points in the sequence. For each point $t_n$ in the sequence, the RL agent takes an action ($a_n \in \{ 0, 1 \})$: either to keep $t_n$ ($a_n = 0$) or to remove it ($a_n = 1$). The RL agent thus learns \textbf{a policy $\pi$ that defines the probability of a point being an outlier}, and if the action sampled from the policy is $1$, the point is removed from the sequence. The generated sequence consists of all the points untouched by the RL agent. In this way, by hopefully keeping only the normal points, it generates a new sequence, which is fed into a discriminator to evaluate how ``real'' it is. The discriminator outputs a reward after the sequence is completed.

\subsection{Discriminator} \label{sec:dis}
The goal of the discriminator is to distinguish between the generated sequences, $S^g_{j}$, obtained from the RL agent, and the real sequences, $S_i$, sampled from the dataset $\mathcal{D}$. As mentioned earlier, since the proportion of corrupted sequences in the dataset should be low, the majority of the samples are clean sequences. The discriminator tries to \textbf{determine whether a given sequence is ``real'' or ``generated''} using a non-linear classifier. To prevent the discriminator from dominating over the RL agent, we also add spectral normalization~\citep{miyato2018spectral} to each layer of the classifier model. The discriminator has the exact same architecture as the generator except without self-attention and layer normalization. However, no weights are shared between them. 

\section{Experiments}
In the experiments, we study, despite only having access to unlabeled and potentially corrupted data, whether \alg\ can still learn outlier detectors without supervision and beat the state-of-the-art approach originally designed for semi-supervised settings (Section~\ref{sec:main_results}). We also add interesting ablation studies and provide explanation of our model and algorithm choices in  Appendix~\ref{sec:ablation}.

\textbf{Datasets} To assess the performance, we use four datasets: two synthetic datasets and two real-world datasets. The synthetic datasets were generated by defining intensity functions for clean data. These intensity functions correspond to either an inhomogeneous \textbf{Poisson} process or a \textbf{Hawkes} process. For real-world datasets, we include the followings: \textbf{MIMIC}, used in prior work in event sequence modelling \citep{du2016recurrent,mei2017neural}, records the admission times of patients in an Intensive Care Unit over a period of 7 years.  \textbf{Taxi}~\citep{taxi}, used in~\citep{xue2022hypro}, tracks taxi pick-up and drop-off events in the New York City.

For all the datasets, we also define a parameter $\beta$, controlling the percentage of clean sequences in the dataset. If $\beta$ is $0.7$, it means that $70\%$ of the sequences are clean. For all our experiments, we use a $\beta$ of $0.8$ unless otherwise mentioned. Outliers are generated by sampling from a Poisson process with a constant intensity function. The specific intensity functions (Appendix~\ref{sec:intensities}) and outlier generation mechanism (Appendix~\ref{sec:outlier_sim}) can be found in the Appendix.

\textbf{Baseline}
As unsupervised event outlier detection without access to clean data has not been previously explored, we adapt the state-of-the-art approach for semi-supervised event outlier detection, PPOD, as our main baseline \citep{liu2021event}, which has demonstrated strong performance when clean data is available. It also leverages the cLSTM architecture but is trained with a negative log-likelihood loss~\citep{mei2017neural} to generate the intensity functions used for scoring points and intervals in the sequences for outliers. As additional baselines, we also include RND, which generates random outlier scores, and LEN, which is based on the inter-event time interval length, from the same paper.

\subsection{Performance of \alg}\label{sec:main_results}
\begin{figure}[ht]
\centering
\begin{subfigure}{0.4\textwidth}
    \includegraphics[width=\linewidth]{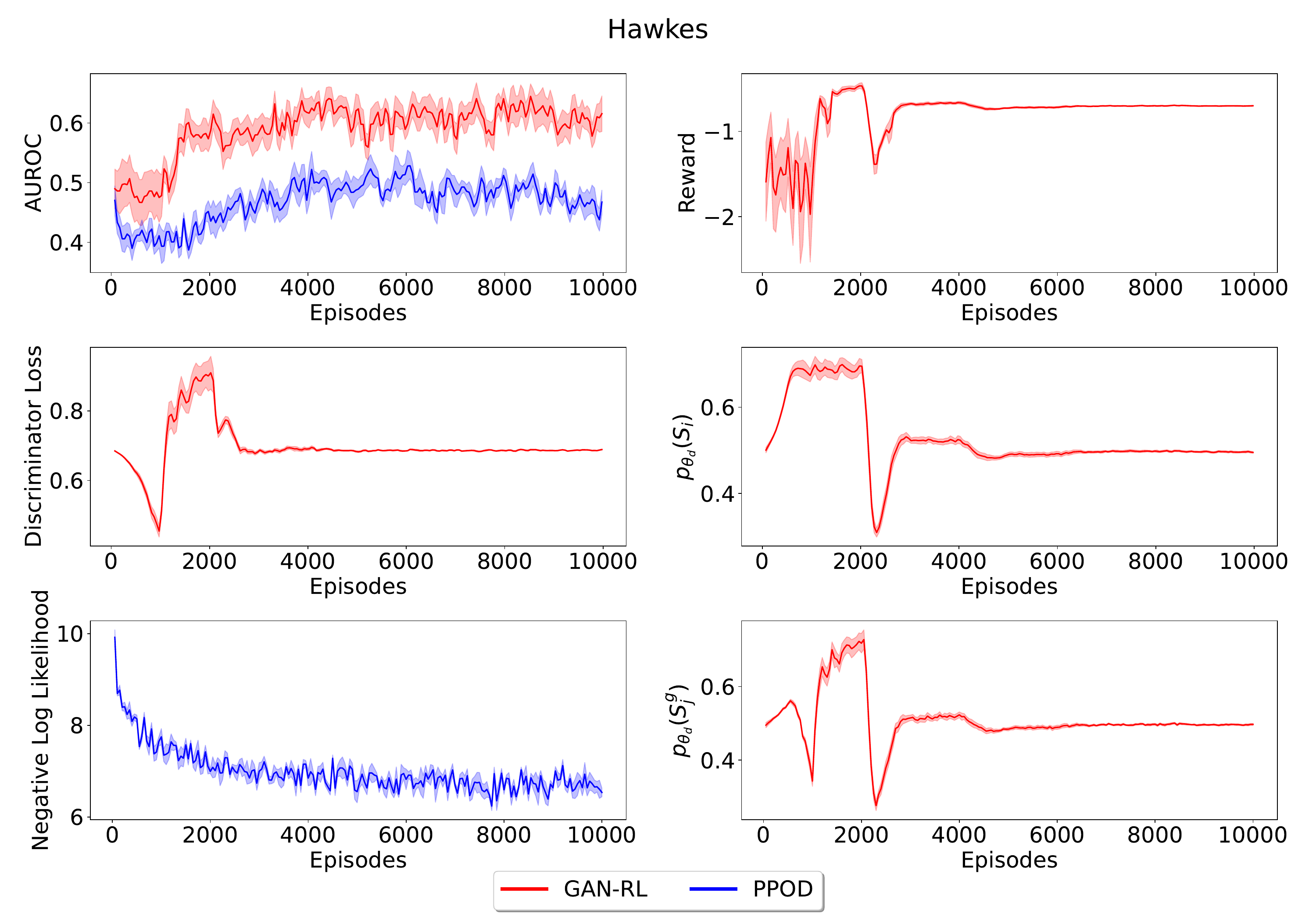}
    \caption{Hawkes}
\end{subfigure}
\begin{subfigure}{0.4\textwidth}
    \includegraphics[width=\linewidth]{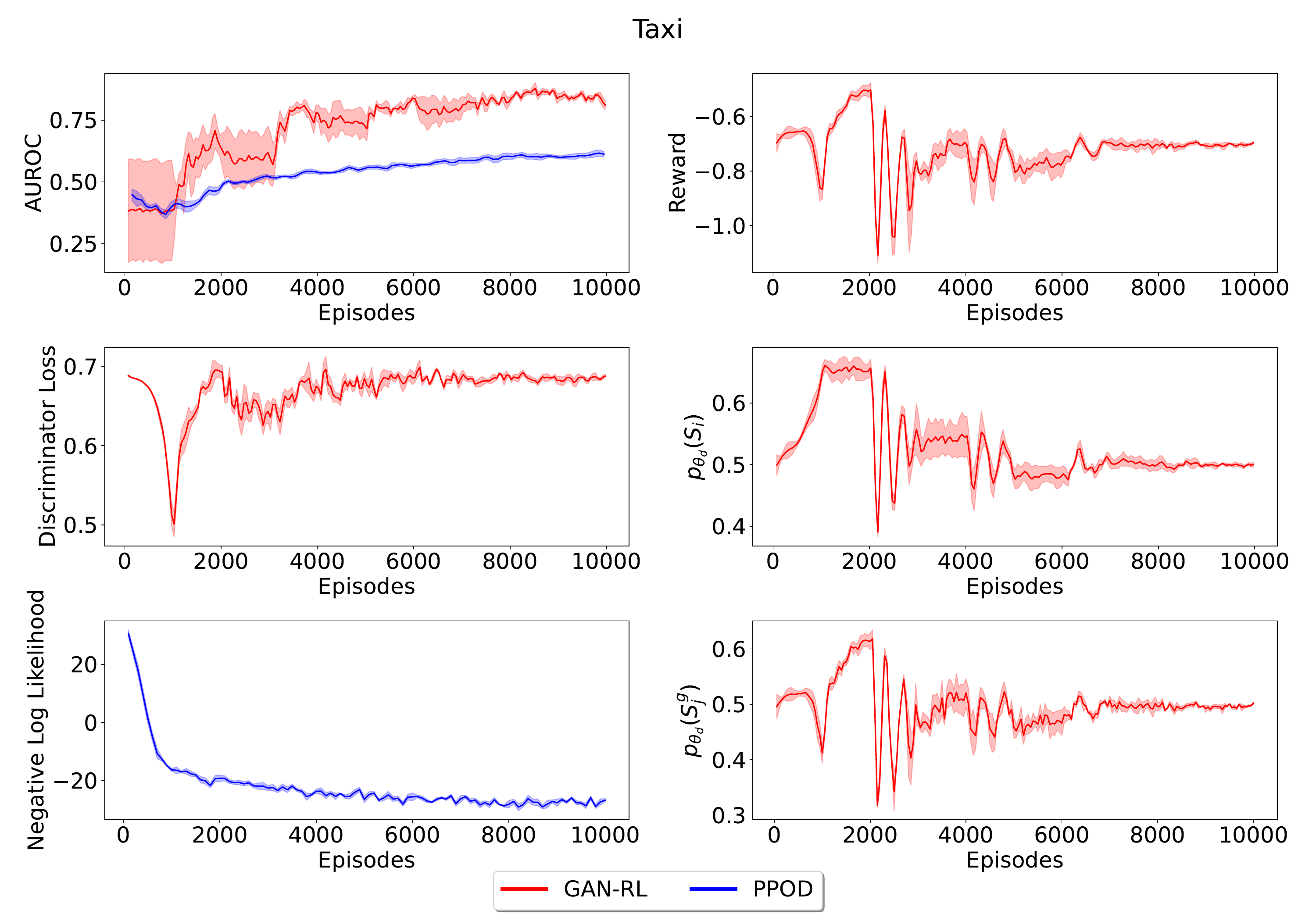}
    \caption{Taxi}
\end{subfigure}
\caption{Training performance of our algorithm on a synthetic and real dataset. We include two more in Appendix~\ref{sec:more_results}. The curves are plotted across 10 independent runs. The shaded regions represent the standard errors.}
\label{fig:comm-syn}
\end{figure}

In Figure~\ref{fig:comm-syn}, we present a comprehensive analysis of our RL agents' performance over a series of 10,000 episodes, each representing a complete sequence. For the synthetic datasets, these sequences are chosen randomly from a pool of 1000 generated sequences for training. Meanwhile, for Taxi, we sample from the same training data splits as used in previous work. To gauge the effectiveness of our method against the baseline method, we employ AUROC (Area Under the Receiver Operating Characteristic Curve) scores, computed based on the ground-truth labels not accessible by any of the methods during training. This metric provides valuable insights into the RL agents' ability to distinguish outliers from the norm. Over 10 independent runs, our results consistently demonstrate that our RL agents excel at outlier detection when compared to the baseline method. Another interesting observation worth mentioning is the probability that the real sequence is classified correctly and the probability that the \emph{generated} is classified as real seems to be around $0.5$ which suggests that the RL agent has done a great job in generating real-looking data by removing outliers. 

\subsection{Results on Test Data}
We also highlight test performance across all the datasets across 10 seeds in Table~\ref{tab:test}. These are computed over 100 test sequences unseen during training. For the real-world datasets, we simply use the test split in the dataset, and for the synthetic datasets, we generate new sequences using seeds different from the ones used in training. These results demonstrate that the improvement in the performance of GAN-RL is applicable to unseen testing data as well, and demonstrates the generalizability of the proposed method.

\begin{table}[ht]
\centering
\caption{Evaluation on test sets.}
\label{tab:test}
\begin{tblr}{
  vlines,
  hlines,
}
Dataset & RND & LEN & PPOD & \alg\\
Poisson & $0.510 \pm 0.029$ & $0.470 \pm 0.023$ & $0.55 \pm 0.02$  & $0.631 \pm 0.03$ \\
Hawkes  & $0.503 \pm 0.032$ & $0.481 \pm 0.031$ & $0.512 \pm 0.01$ & $0.610 \pm 0.03$ \\
MIMIC   & $0.495 \pm 0.014$ & $0.337 \pm 0.024$ & $0.583 \pm 0.02$ & $0.778 \pm 0.12$ \\
Taxi    & $0.503 \pm 0.007$ & $0.587 \pm 0.005$ & $0.548 \pm 0.01$ & $0.647 \pm 0.03$
\end{tblr}
\end{table}

\section{Conclusion}

In this work, we developed a novel \emph{unsupervised} event outlier detection framework based on ideas from GANs and RL. RL-based generators are learned to correct outliers in the unlabeled dataset through GAN-based training and then applied to unseen sequences to detect event outliers online. We evaluated our method on both synthetic and real-world datasets with simulated outliers. Compared with the state-of-the-art semi-supervised approaches, our method shows similar or better detection accuracy in all the experiments.

\bibliography{refs1,refs2}
\bibliographystyle{iclr2024_conference}

\appendix
\section{Related Work}
\label{sec:related_work}
\paragraph{Generative model based outlier detection}
Although outlier detection methods using deep generative models are not new (e.g.,~\citep{akcay2019ganomaly,schlegl2017unsupervised,li2019mad-gan,zhu2023sequential}), none of them detect abnormal occurrences and absence of events, and almost all of them take a semi-supervised approach, assuming availability of training data consisting of samples from a single class (usually normal data but sometimes abnormal data). Note that some authors~\citep{schlegl2017unsupervised,li2019mad-gan} just use a different terminology. In contrast, our method detects event outliers in continuous time and does it in a completely unsupervised setting, where the training event sequences contain both normal and abnormal occurrences and absence of events. Once trained, the models can be used for detecting outliers online on unseen data.

\paragraph{Deep generative models for event sequences}

Most deep learning models for event sequence data are based on combining deep architectures (such as RNNs~\citep{du2016recurrent,mei2017neural} and attention-based~\citep{zhang2020selfattentive,zuo2020transformer} with TPPs, and learned by maximizing the likelihood. Different from these models, \citet{xiao2017wasserstein} developed a Wasserstein distance for TPPs and a Wasserstein GAN to generate samples from the learned TPP. \citet{li2018learning} propose to use reinforcement learning to learn a generative model utilizing inverse reinforcement learning to learn the reward from the training data. Our work is also inspired by ideas from GANs and reinforcement learning, but our goal is outlier detection instead of sequence generation, which results in different model architectures and algorithms for learning and inference.

\paragraph{Outlier detection for event sequences}

\citet{liu2021event} aim to detect event outliers as in our work, but their approach requires training a point process model to learn the distribution of the data, which, in theory, would require the training data to be clean without any outliers. In contrast, our method does not require clean data and can learn directly on polluted data in a completely unsupervised fashion. \citet{zhang2021learning} develop a greedy algorithm to separate exogenous events from an event sequence, which can be considered as a special case of unexpected event occurrences. Similarly \citet{mei2019imputing} propose a particle-smoothing algorithm for imputing missing events in event sequences. These methods focus on offline data processing and analysis, while our work focuses on unsupervised learning of models for online outlier detection. \citet{zhu2023sequential} propose a GAN-based approach to detecting anomalous \emph{sequences} assuming the availability of only anomalous data, while we focus on detecting anomalous occurrences and absence of \emph{events} assuming the unlabeled data contain both normal and abnormal events. Similarly \cite{shchur2021detectinga} propose a new statistic for goodness-of-fit testing and detecting anomalous event \emph{sequences} instead of events.

\section{Algorithm Details}\label{sec:alg}
Algorithm~\ref{alg:uod-rl} describes how the generator and discriminator are trained and what the inputs and outputs of the models look like. One additional thing to note is that, while each sequence is treated as an episode for the RL agent, the generator is trained only when an episode has been completed i.e. we always wait for the sequence to get completed before updating the generator. The same is true for the discriminator because it would need complete sequences as inputs anyway. Another crucial thing to keep in mind is that although this entire network can be trained end-to-end, we opt for an \emph{iterative} training regime in the spirit of alternating gradient descent in the GAN literature \citep{mescheder2018training}. The difference is that our generator is trained by PPO instead of gradient descent. That means for the first few episodes only the discriminator is trained, and then only the generator, and so on. This is to ensure that the training dynamics are sufficiently smooth and that both the discriminator and the generator can learn gradually. In Section \ref{sec:trai_dynamics}, we also illustrate the entire model architecture and provide intuition on the training dynamics with more details on the training framework.
\begin{algorithm} 
\caption{\alg}\label{alg:uod-rl}
\begin{algorithmic}[1]
\Require Unlabeled Event Sequences $\mathcal{D} = \{S_i\}_{i=1}^I$, \texttt{Generator} ($\theta_g$), \texttt{Discriminator} ($\theta_d$), Update Frequency $F$, Number of Episodes $K$
\For{$k=0 \dots K-1$}
    \State Sample $S_{j} = \{t_n\}_{n=1}^{N_j} \sim \mathcal{D}$
    \State $\{\phi^{j}_n\}_{n=1}^{N_j} = \texttt{Encoder}_{\theta_g}(S_j)$
    \For{$n=1 \dots N_j$}
        \State $a_n \leftarrow \pi_{\theta_g}(\phi^{j}_n)$
        \State $r_n = 0$
    \EndFor
    \State $S^{g}_j = \texttt{generate}(S_j, \{a_n\}_{n=1}^{N_j})$ \Comment{remove points}
    \If{$k \mod F < F / 2$}
    
    \State Sample $S_i \sim \mathcal{D}$ and compute $p_{\theta_d}(S_i)$ and $p_{\theta_d}(S_j^g)$
    \State Update $\theta_d$ using Cross Entropy Loss
    \Else
    \State Compute $r_{N_j} =  p_{\theta_d}(S_j^g)$
    \State Update $\theta_g$ using RL Loss
    \EndIf
\EndFor
\end{algorithmic}
\end{algorithm}
\subsection{Training Dynamics}\label{sec:trai_dynamics}
\begin{figure}[h]
    \centering
    \includegraphics[width=\linewidth]{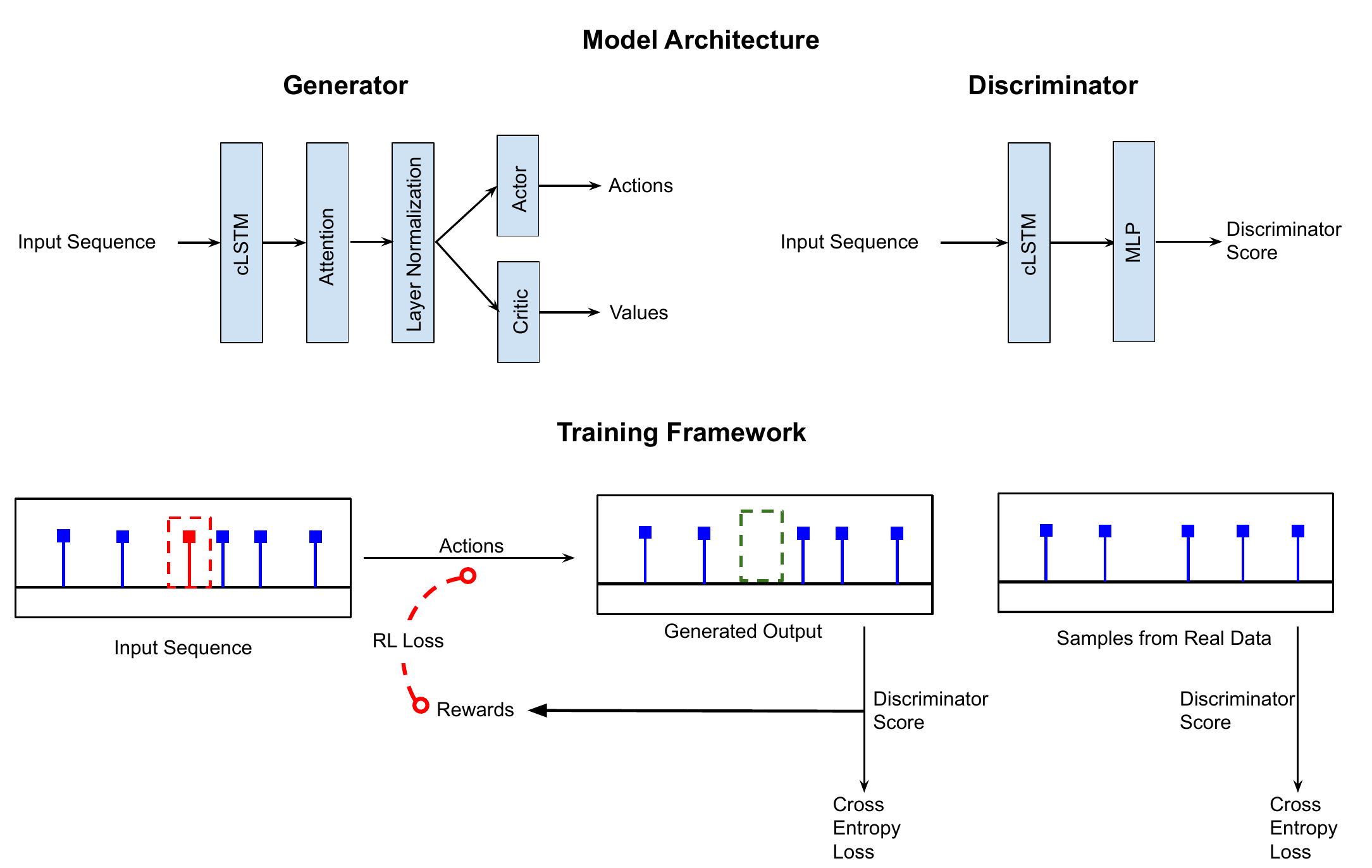}
    \caption{Model Architectures and Training Framework}
    \label{fig:training}
\end{figure}
In Figure~\ref{fig:training}, we illustrate the individual model architectures for both the generator and discriminator. Both the generator and the discriminator encode sequences with separate cLSTMs. The generator additionally has a causal attention layer, followed by layer normalization. These encodings are then fed to the actor and critic which output actions and the value functions respectively. The discriminator architecture is much simpler, and the cLSTM outputs are just fed to a neural network to output the scores for each of the individual sequences.

Figure~\ref{fig:training} also describes the training framework and how the RL and Discriminator interact with each other. The RL agent generates new sequences from inputs by taking actions. The discriminator score is fed as the reward to the input sequence. The discrimination score is the probability that the input sequence is a 'real' sequence, so the RL agent should try to improve this score, hence this score is an ideal candidate to be chosen as the reward. The discriminator is trained by its separate cross-entropy loss.

At the core of the training methodology for this framework lies the concept of leveraging a single scalar reward function for each sequential dataset. During discriminator training, a generated sequence has a target label 0 and a real sequence 1. In parallel, the RL agent takes on a distinct objective: to enhance the quality of the generated sequences, striving to render them as devoid of outliers as possible, thereby aligning them closely with the characteristics of clean sequences. The incentive guiding the RL agent's actions is grounded in the discriminator's output for classifying sequences as 0. This output is repurposed as a reward signal that propels the RL agent to optimize its approach.
This interplay between the RL agent and the discriminator creates a dynamic wherein the generated sequences evolve to closely mimic the clean sequences, blurring the boundaries between the two gradually. 

\section{Datasets}\label{sec:datasets}
\subsection{Intensities for Synthetic Datasets}\label{sec:intensities}
The synthetic datasets are each characterized by their own intensity functions. For simulating these Point Processes, we use Tick~\citep{tick}. The details are as follows:
\begin{itemize}
    \item \textbf{Poisson}: The intensity function, $\lambda(t) = 1 + \sin(2t)$.
    \item \textbf{Hawkes}: we use a kernel with a sum of $U$ exponential decays with intensities $\alpha = [0.01,0.02,0.01]$ and decays $\beta = [1.0, 3.0, 7.0]$. The intensity function is defined as:
    $$ \lambda(t)=\sum_{u=1}^U \alpha_u \beta_u \exp \left(-\beta_u t\right) 1_{t>0} $$
\end{itemize}

\subsection{Outlier Simulation}\label{sec:outlier_sim}
We use a Poisson process with a constant intensity function ($\alpha$) to generate outliers, and then we merge them with the clean sequence to create a corrupted sequence. The value of the intensity function is defined based on the type of dataset being used.

As a rule of thumb, we choose $\alpha$ such that the number of outliers in a sequence is around $20-30\%$ of the average clean sequence length. The exact values for every dataset are in Table \ref{tab:hyps-comm}.

\section{Additional Training Curves on Poisson and MIMIC}\label{sec:more_results}
\begin{figure}[ht]
\centering
\begin{subfigure}{0.45\textwidth}
    \includegraphics[width=\linewidth]{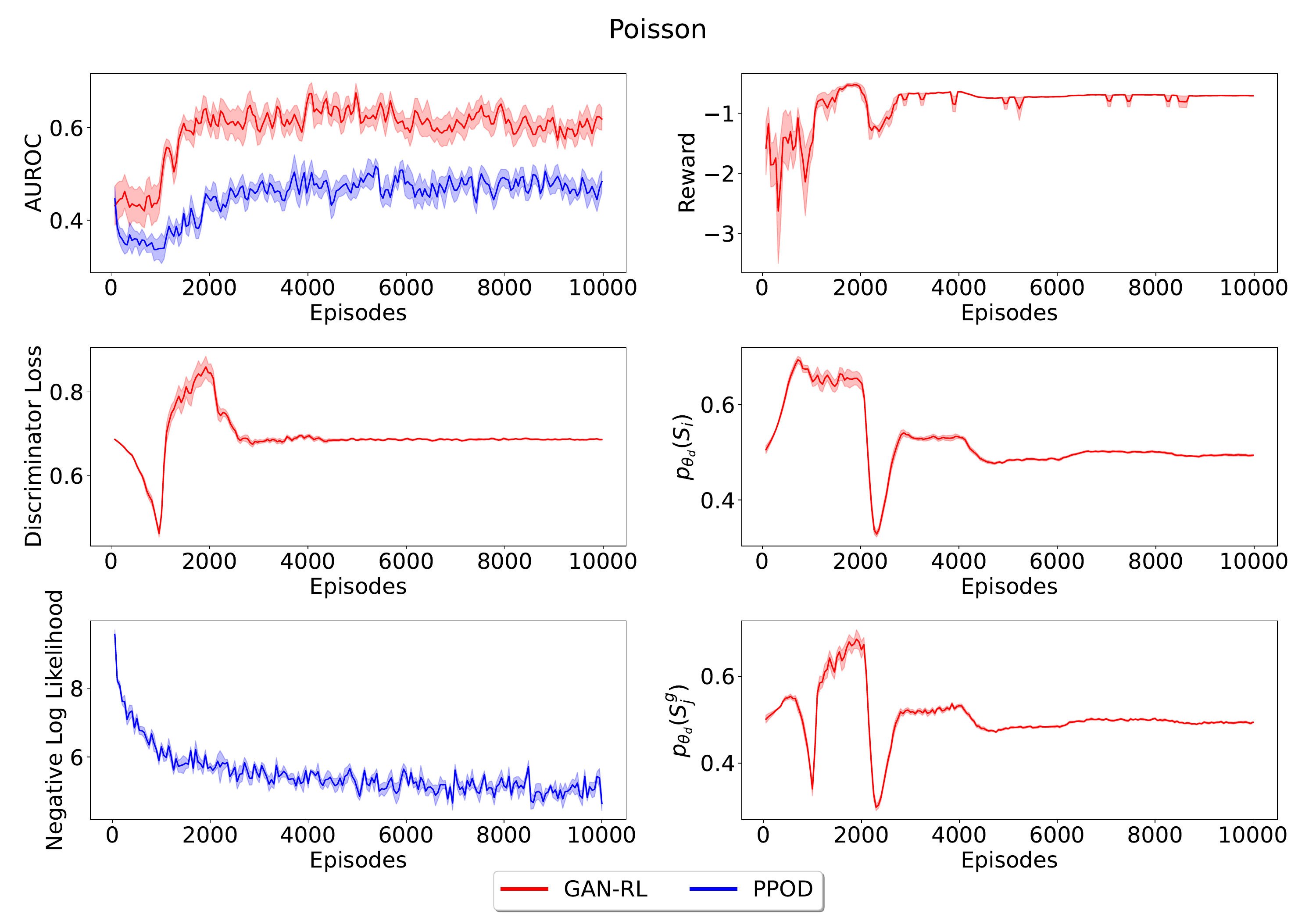}
    \caption{Poisson}
\end{subfigure}
\begin{subfigure}{0.45\textwidth}
    \includegraphics[width=\linewidth]{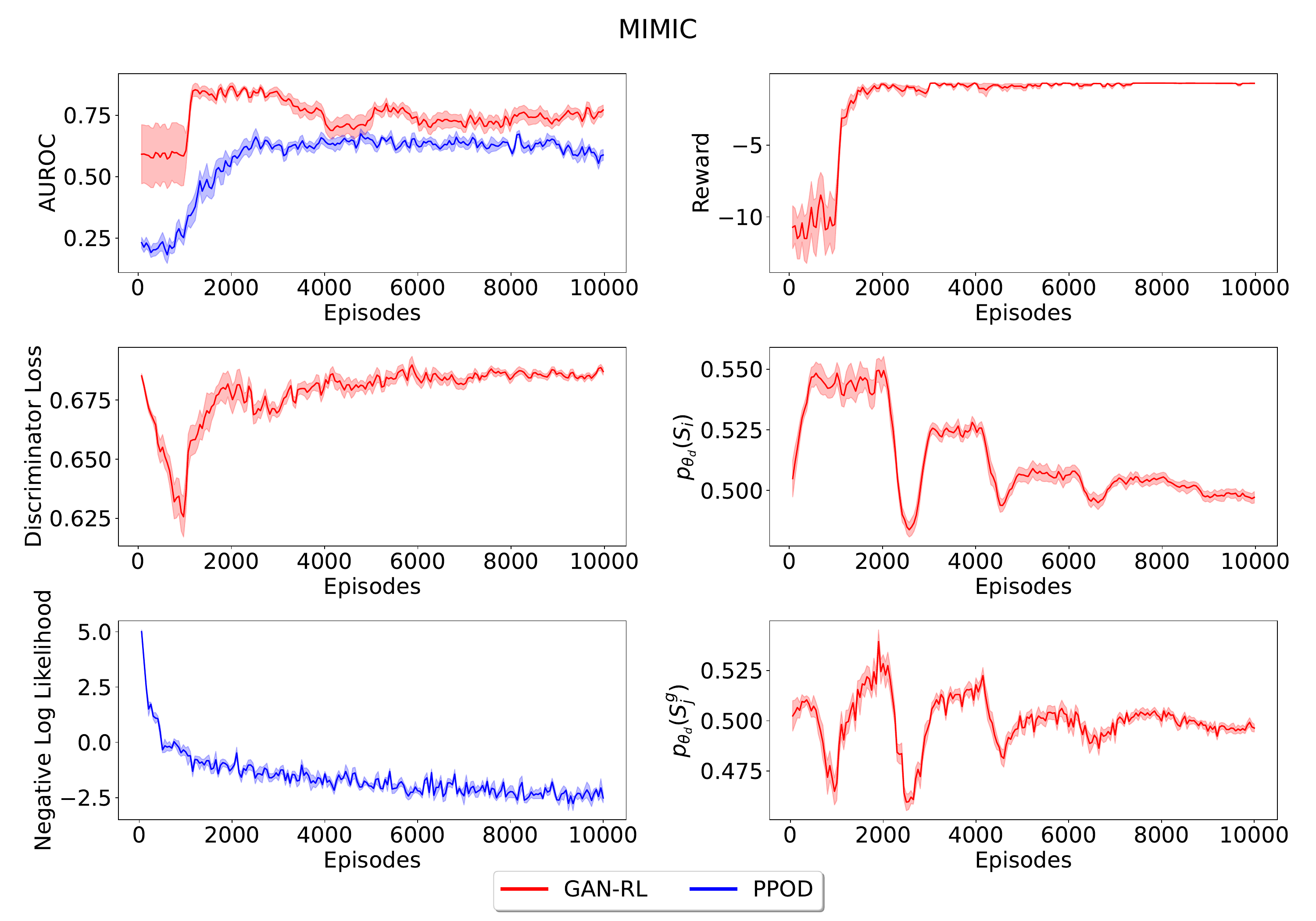}
    \caption{MIMIC}
\end{subfigure}
\caption{Training performance of our algorithm on one synthetic and real dataset. The curves are plotted across 10 independent runs. The shaded regions represent the standard errors.}
\label{fig:comm-2}
\end{figure}

\section{Ablation Studies}\label{sec:ablation}
\subsection{Sensitivity to the Amount of Corruption in the Data}\label{sec:beta}
\begin{figure}[ht]
\begin{subfigure}{0.45\textwidth}
    \includegraphics[width=\linewidth]{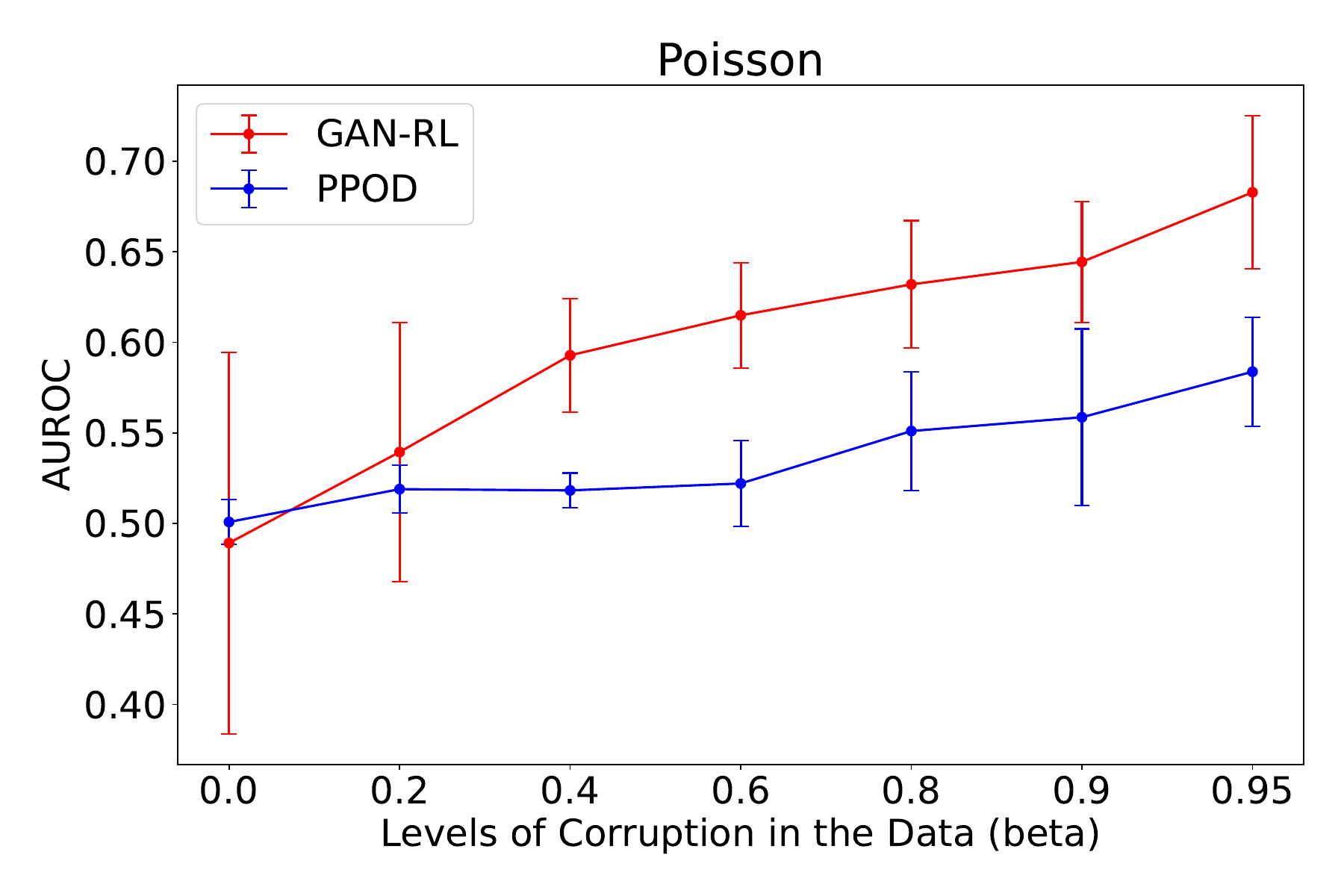}
\end{subfigure}
\begin{subfigure}{0.45\textwidth}
    \includegraphics[width=\linewidth]{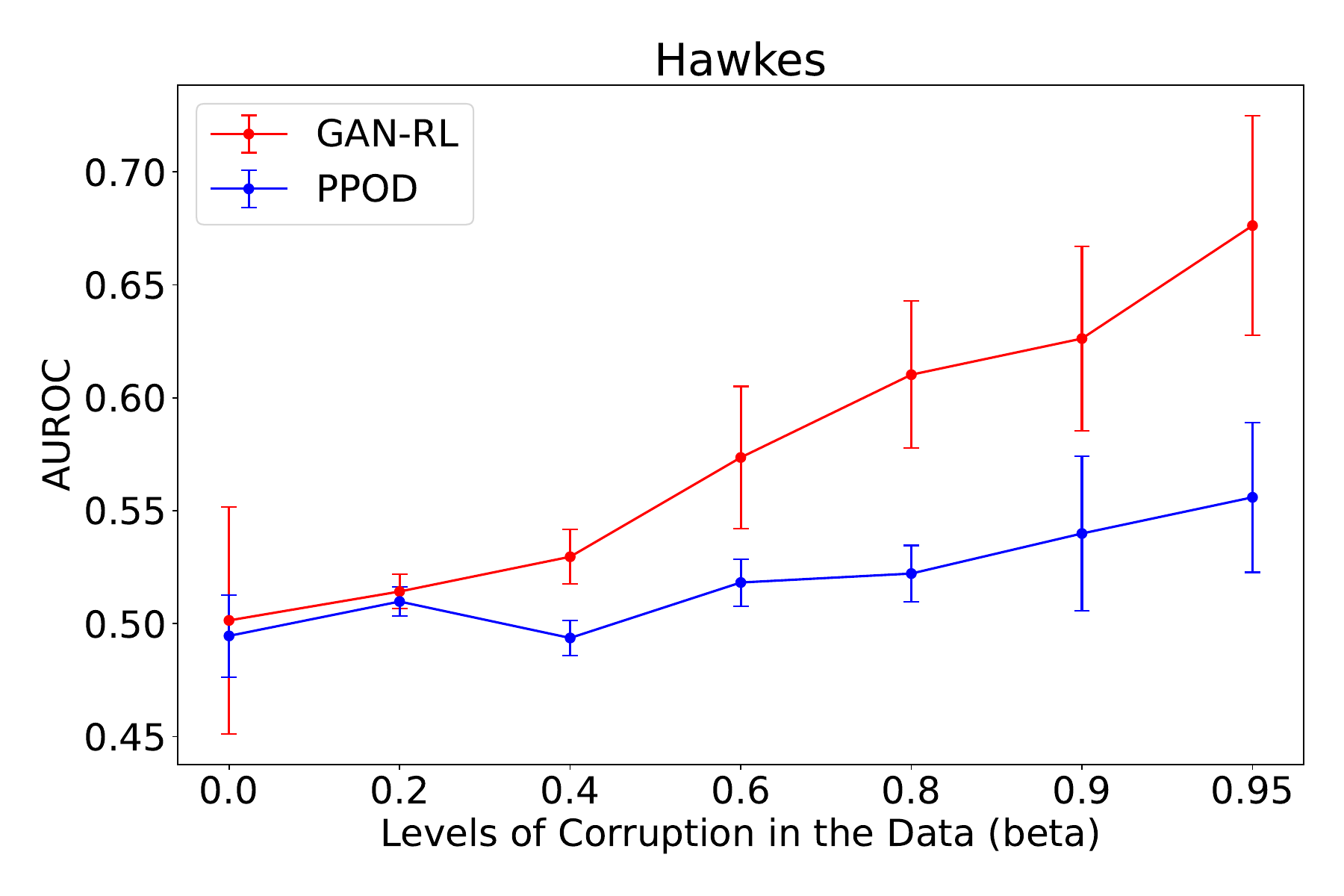}
\end{subfigure}
\newline
\begin{subfigure}{0.45\textwidth}
    \includegraphics[width=\linewidth]{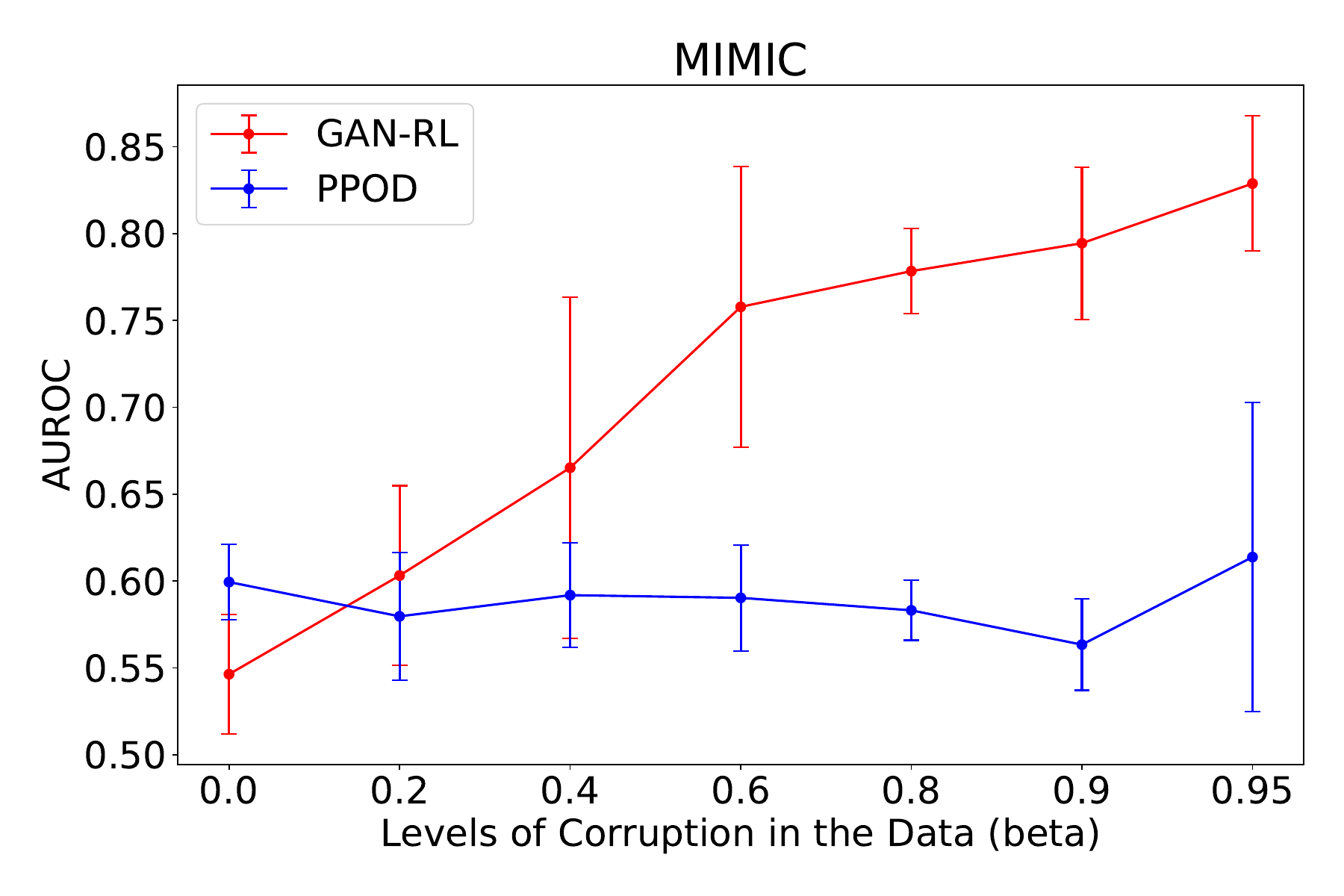}
\end{subfigure}
\begin{subfigure}{0.45\textwidth}
    \includegraphics[width=\linewidth]{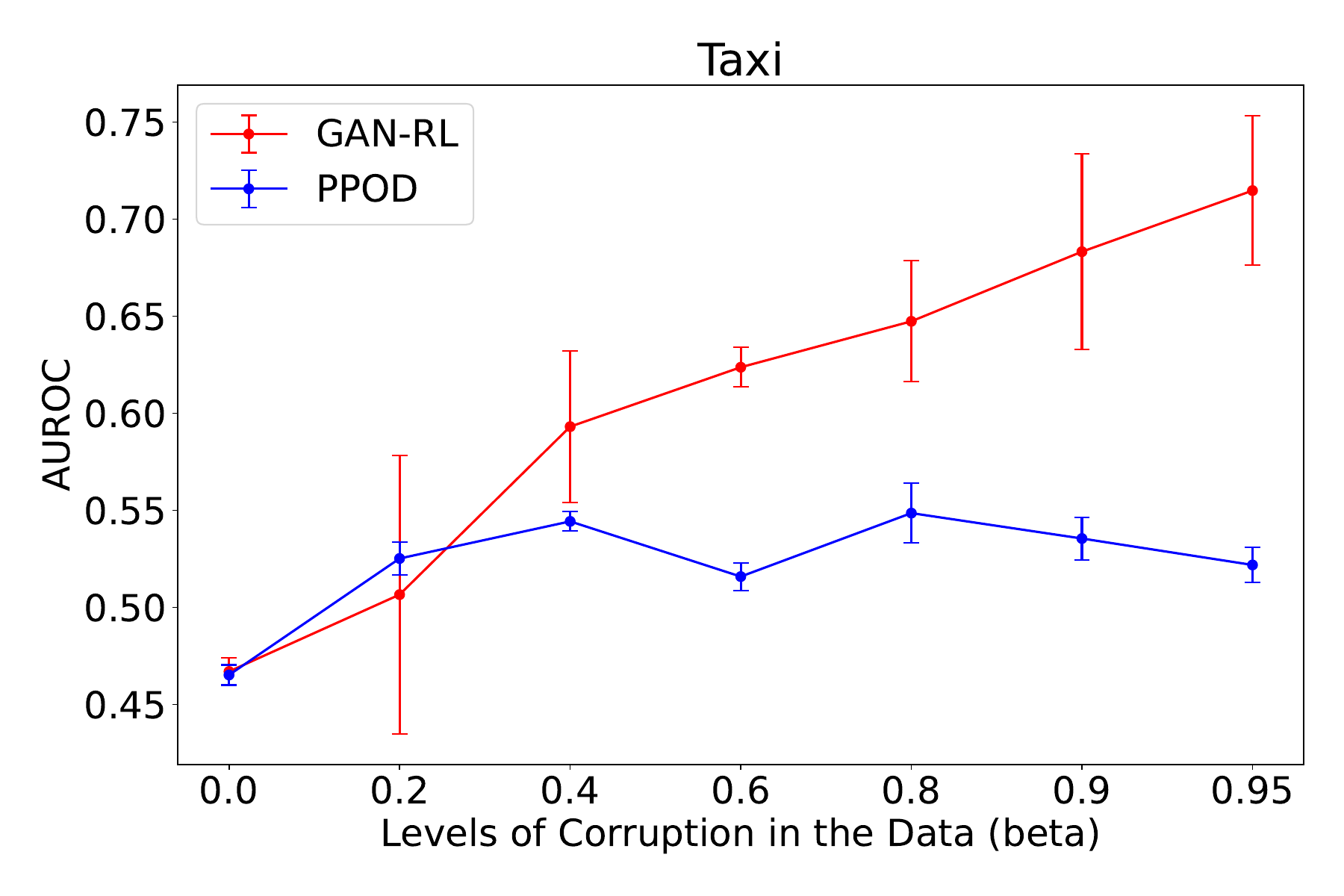}
\end{subfigure}
\caption{Sensitivity to corruption in the data.}
\label{fig:beta}
\end{figure}

The unsupervised nature of the problem setting necessitates the presence of \emph{some} normal sequences (or sequence segments if we split complete sequences into shorter segments) in the dataset, and this usually should not be an issue as the majority of the data should be clean by definition of outliers. However, if that is not the case we expect \alg\ to fail because the discriminator would perceive noisy sequences as real.  To study the effect of the ``cleanness'' of the datasets, we plot in Figure~\ref{fig:beta} the average performance on 100 test sequences. The $X$-axis is the parameter $\beta$ and increasing $\beta$ means fewer corrupted sequences. As $\beta$ increases, we notice an improvement in the performance of \alg\ as the discriminator gets more real sequences and is able to distinguish them better from generated data. However, when $\beta = 0.0$, we notice that \alg\ gets an AUROC of around $0.5$ suggesting it is purely random. These results demonstrate how much we can relax the assumption of most of the data being clean. From Fig.~\ref{fig:beta}, we see GAN-RL can perform well even with 60\% of clean data across all the datasets.

\subsection{Importance of Attention}\label{sec:attn}
\begin{table}[ht]
\centering
\caption{Comparison of asymptotic training performance with and without attention.}
\label{tab:attn}
\begin{tblr}{
  vlines,
  hlines,
}
Dataset &
  \alg\ w/o Attention & \alg\  &
  PPOD & PPOD with Attention \\ 
Poisson & 0.542 $\pm$ 0.02 & 0.651 $\pm$ 0.03 & 0.506 $\pm$ 0.01 & 0.503 $\pm$ 0.03 \\ 
Hawkes  & 0.522 $\pm$ 0.01 & 0.654 $\pm$ 0.02 & 0.515 $\pm$ 0.04 & 0.523 $\pm$ 0.08  \\ 
MIMIC    & 0.592 $\pm$ 0.05 & 0.601 $\pm$ 0.08 & 0.593 $\pm$ 0.06  & 0.582 $\pm$ 0.04  \\
Taxi    & 0.795 $\pm$ 0.01 & 0.832 $\pm$ 0.02 & 0.690 $\pm$ 0.01  & 0.710 $\pm$ 0.01  \\ 
\end{tblr}
\end{table}

The addition of the attention layer was really crucial to the good performance of \alg. We believe this is because the generator needs to be able to focus on particular time points in the past before being able to make decisions on outliers and this is where adding attention can help. We run an ablation study where we remove the attention layer from the encoder and plot the training performance across the last $10\%$ episodes. 
We do notice a drop in performance of \alg\ without attention in Table~\ref{tab:attn}. It is interesting to note that, although removing attention results in worse performance of \alg\ in general, the amount of decrease can vary from datasets to datasets. Importantly, even without attention, \alg\ can still outperform the baseline across all datasets, especially on Taxi, suggesting there are merits to our overall framework. Meanwhile, just adding an attention layer on top of cLSTM to the baseline does not improve performance much, if at all.

\subsection{Utility of GANs}
\begin{figure}[ht]
\begin{subfigure}{0.45\textwidth}
    \includegraphics[width=\linewidth]{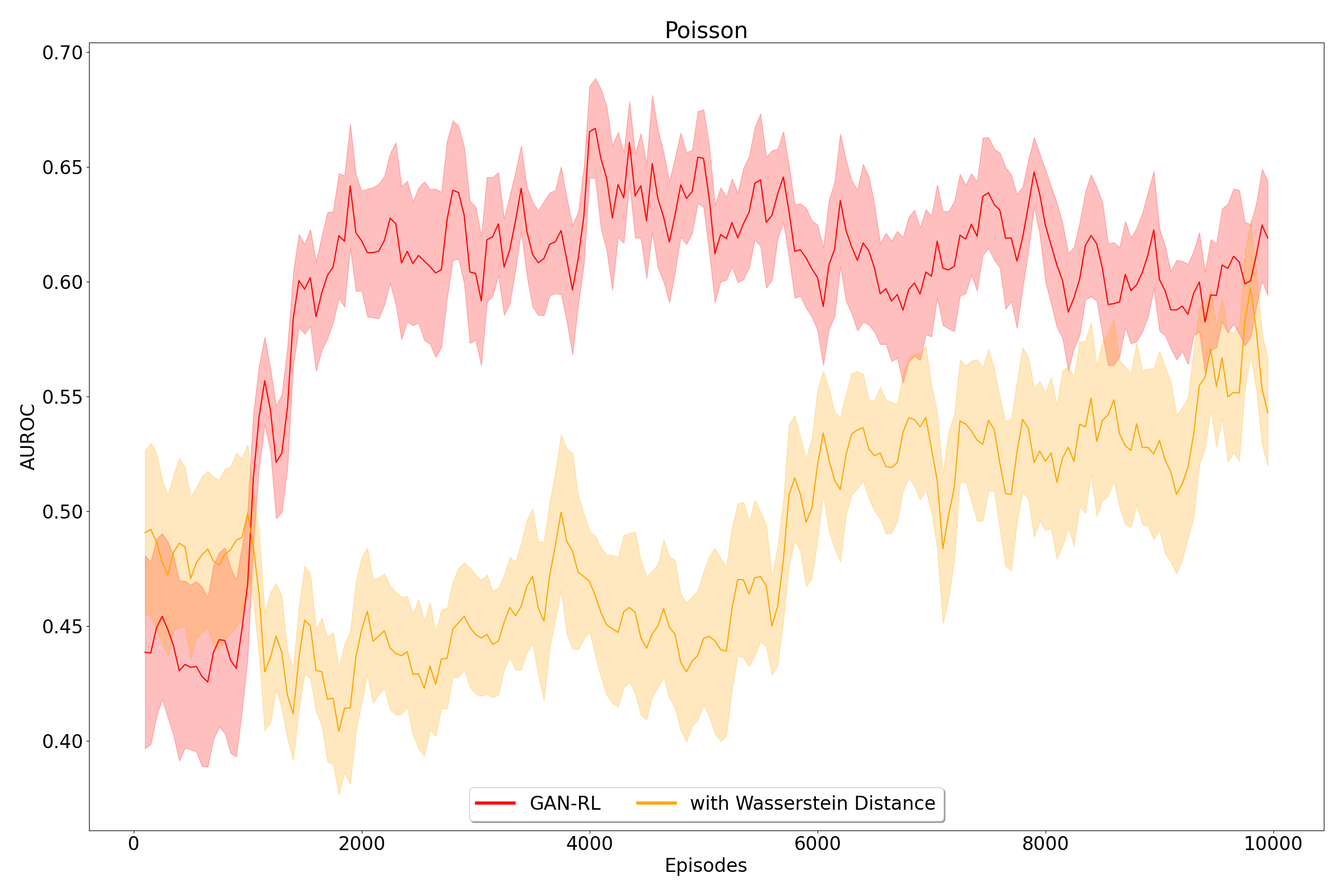}
    \caption{Poisson}
\end{subfigure}
\begin{subfigure}{0.45\textwidth}
    \includegraphics[width=\linewidth]{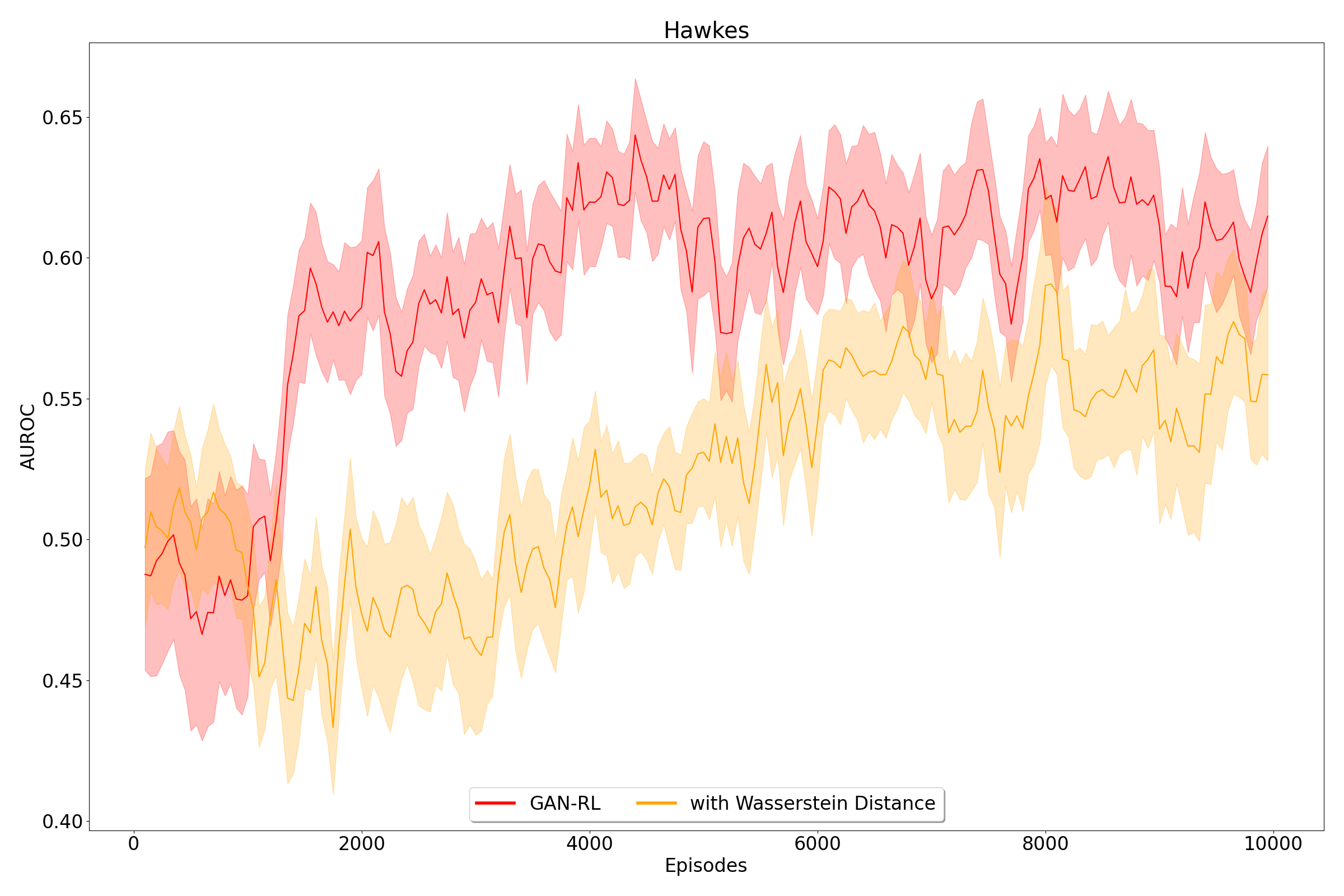}
    \caption{Hawkes}
\end{subfigure}
\caption{Comparison of GAN-RL with a static reward function that operates on the Wasserstein Distance between the generated and real sequences.These results are across 10 independent runs.}
\label{fig:wd}
\end{figure}
We look at a simple RL agent that takes actions to generate clean sequences. Instead of using a Discriminator to evaluate the quality of the generated sequence, here, we use the Wasserstein distance between the real and generated sequence as a reward signal. So, essentially, we have a non-learnable reward signal. Figure~\ref{fig:wd} portrays GAN-RL along with new variation without the discriminator. From the difference in AUROC during training, we can conclude that having a learnable discriminator is very crucial to our method. This further justifies our decision to introduce a GAN based framework for outlier detection.

\subsection{Learning End-to-End}
\begin{figure}[ht]
\begin{subfigure}{0.45\textwidth}
    \includegraphics[width=\linewidth]{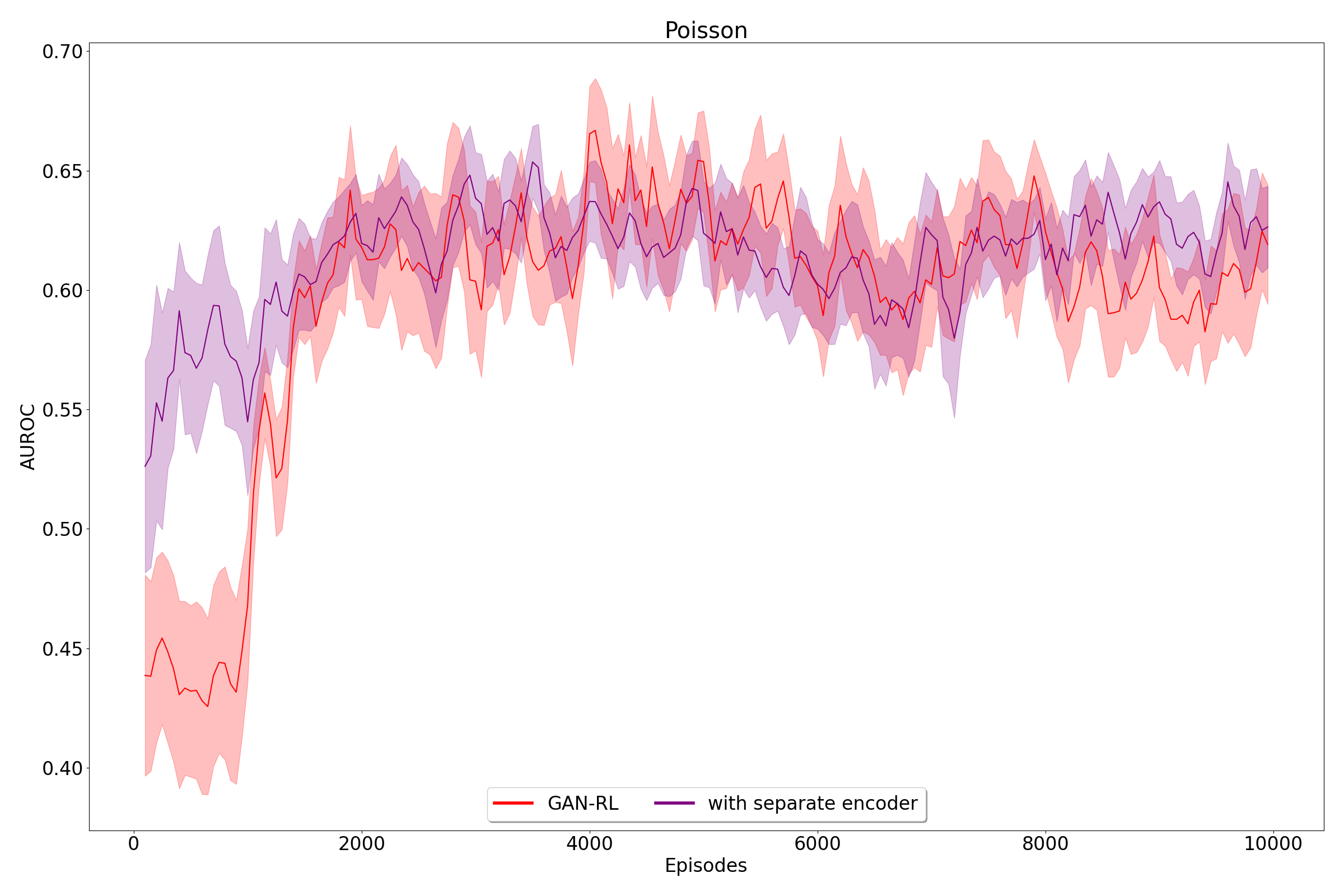}
    \caption{Poisson}
\end{subfigure}
\begin{subfigure}{0.45\textwidth}
    \includegraphics[width=\linewidth]{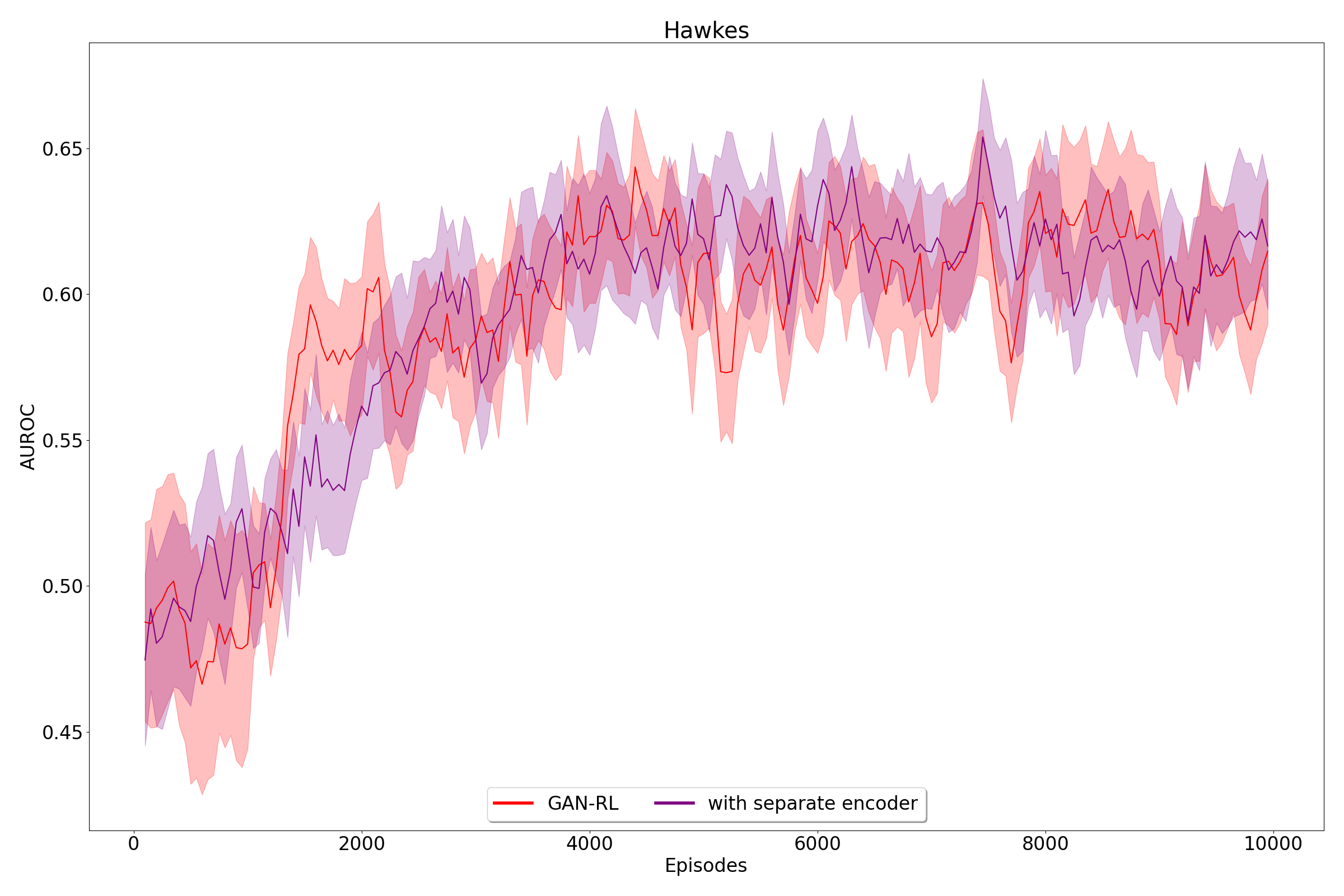}
    \caption{Hawkes}
\end{subfigure}
\caption{Comparison of GAN-RL with a variant of GAN-RL that uses a separately trained encoder cLSTM that is trained on the Negative Log Likelihood loss similar to PPOD. These results across 10 independent runs.}
\label{fig:nll}
\end{figure}

For the second ablation, we try to evaluate if we lose any performance by training the encoder directly with the generator or discriminator losses. In Figure~\ref{fig:nll}, we show a variation of GAN-RL with a separate encoder that is trained exactly like the PPOD baseline. Our method is applied on top of it, and hence the method is not end-to-end. When we try to compare it with our method, we see that we do not give up much performance when compared with the method that has a separately trained encoder.

\section{Hyper-Parameter Sensitivity}
\begin{figure}[ht]
\begin{subfigure}{0.32\textwidth}
    \includegraphics[width=\linewidth]{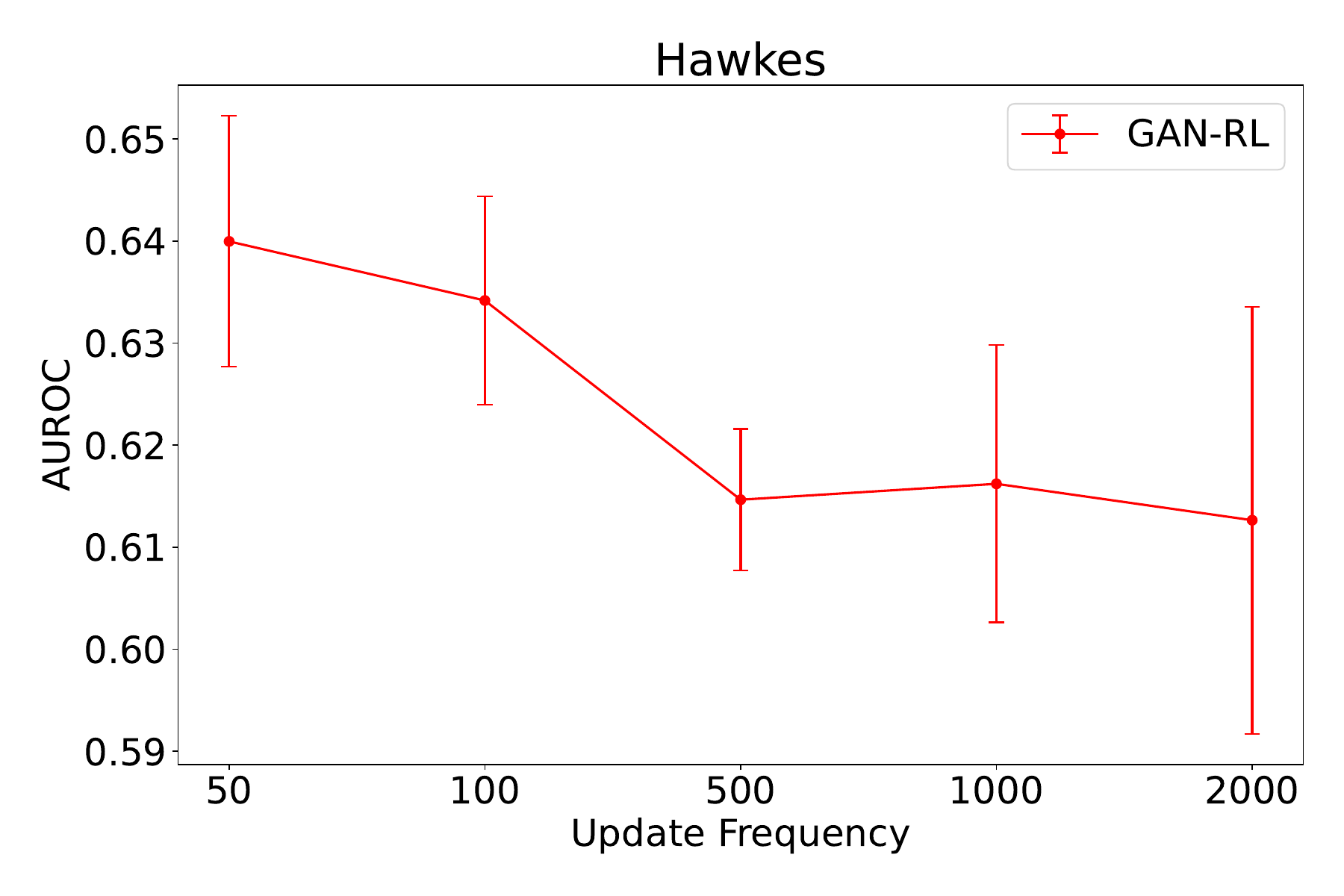}
    \caption{Update Frequency}
\end{subfigure}
\begin{subfigure}{0.32\textwidth}
    \includegraphics[width=\linewidth]{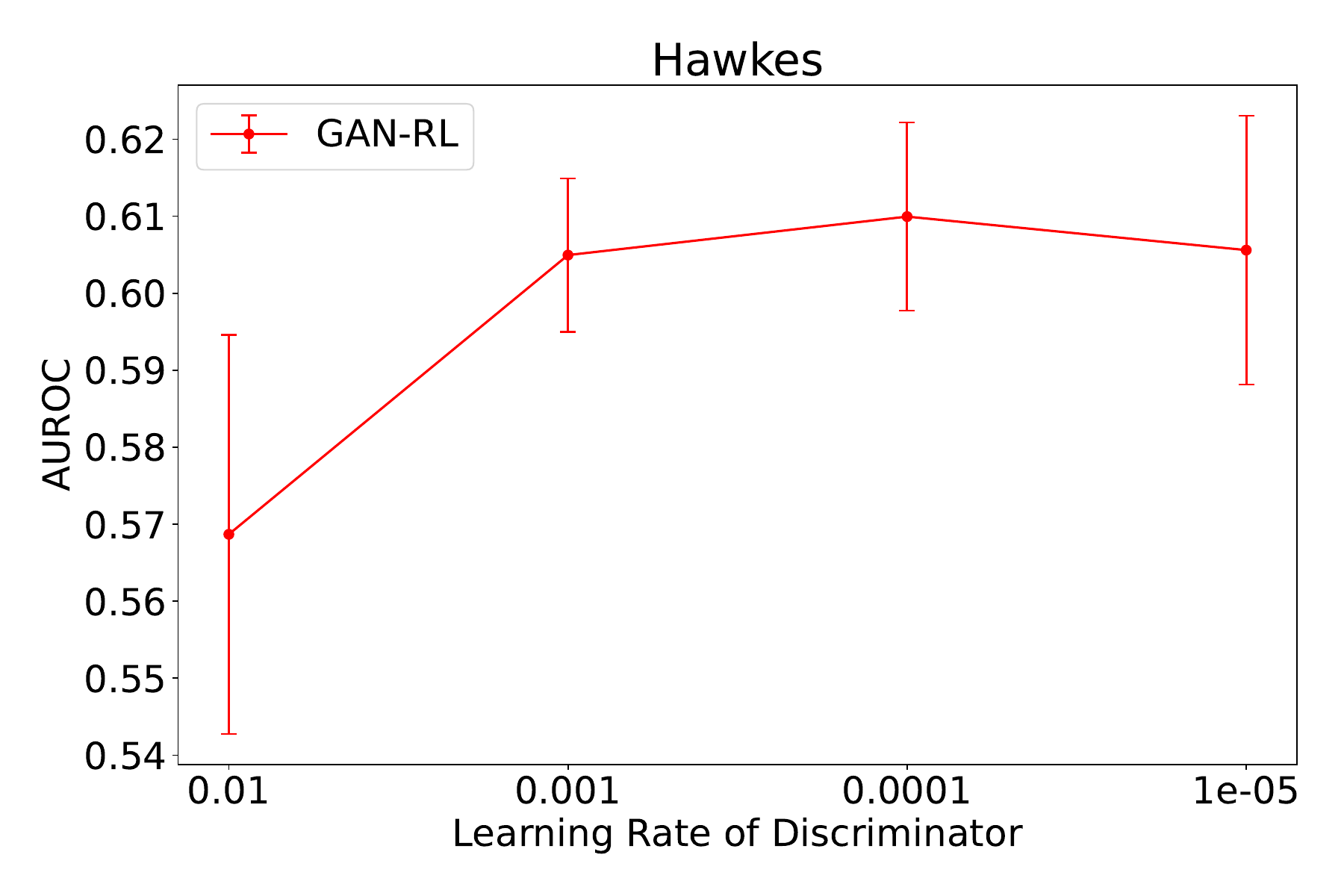}
    \caption{Discriminator Learning Rate}
\end{subfigure}
\begin{subfigure}{0.32\textwidth}
    \includegraphics[width=\linewidth]{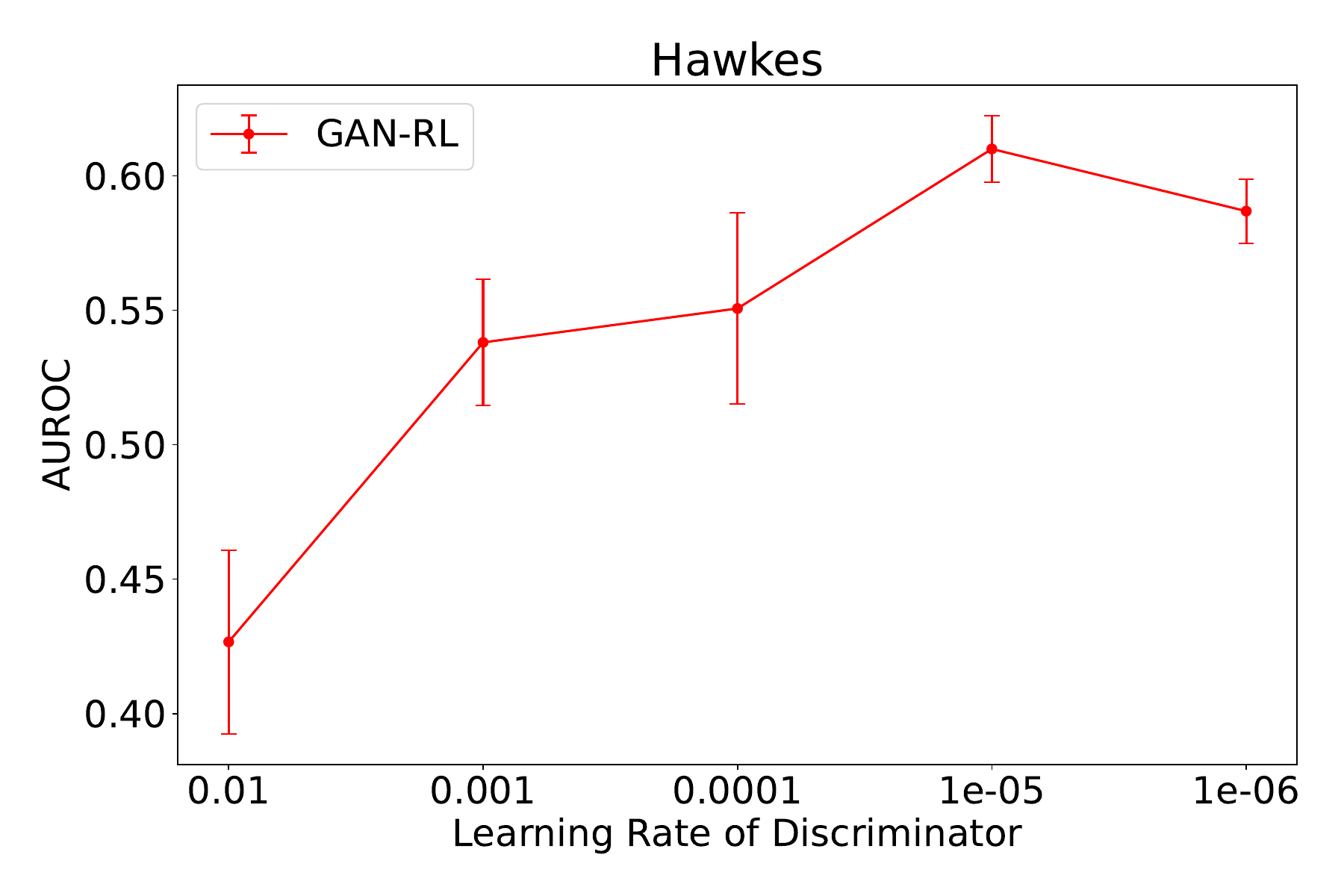}
    \caption{Generator Learning Rate}
\end{subfigure}
\caption{Sensitivity curves across 10 runs for Hawkes dataset.}
\label{fig:sensi}
\end{figure}

In this section, we evaluate the sensitivity to the key hyper-parameters of our method. We believe there are three key hyper-parameters that dictate the learning dynamics. The first is the update frequency which is the number of episodes after which the generator or the discriminator training is switched. This is highlighted in Fig~\ref{fig:sensi} (a), where we see relatively good performance across different values, suggesting our algorithm is quite robust to this hyper-parameter.

Additionally, the learning rate of the generator and the discriminator also influence the training dynamics and these are shown in Fig~\ref{fig:sensi} (a) \& (b). Here, we notice a very similar trend. Higher values of the learning rate of both the generator and discriminator are quite bad, and we believe this is because of the inter-dependency of the generator and discriminator. For higher values, the discrimiantor for example can take bigger gradient steps in the wrong direction thus messing up the reward structure for the RL agent and this can lead to bad performance overall.

\newpage
\section{Hyperparameters}\label{sec:hyps}
For all the experiments, the hyper-parameters were tuned on final AUROC scores on the training set.
\begin{longtblr}[
  caption = {Hyperparameters for all the experiments },
  label = {tab:hyps-comm},
]{
  width = \linewidth,
  colspec = {Q[204]Q[383]Q[352]},
  hlines,
  vlines,
}
Dataset & Generator & Discriminator\\
{Maximum Time Length=10\\Poisson: $\alpha$=0.5\\ Hawkes: $\alpha$=0.5\\MIMIC: $\alpha$=0.1\\Taxi: $\alpha$=0.3\\Seeds:\\Training: $[100,...,109]$\\ Testing: $[1000]$} & {cLSTM hidden size=64\\Self Attention Layer\\Layer Normalization\\learning rate for encoder = 0.001\\\\Actor Arch:\\Linear(64, 64),\\Tanh(),\\Linear(64, 64),\\Tanh(),\\Linear(64, 2),\\Softmax()\\\\Critic Arch:\\Linear(64, 64),\\Tanh(),\\Linear(64, 64),\\Tanh(),\\Linear(64, 2),\\\\update policy every 10 sequences\\update policy for 10 epochs in one update\\clip parameter for PPO, $\epsilon = 0.2$\\$c_1=0.5$, $c_2=0.01$\\discount factor = 0.99\\learning rate for actor network = 0.00001\\learning rate for critic network =~~0.00001} & {cLSTM hidden size=64\\learning rate for encoder = 0.001\\\\Arch:\\Linear(64, 64)\\Tanh(),\\Linear(64, 1)\\Linear Layers have \\Spectral Normalization\\\\update discriminator every 50 sequences\\learning rate for discriminator = 0.001\\Number of Episodes=10000\\Update Frequency=1000\\}\\
\end{longtblr}

\end{document}